\documentclass[runningheads]{llncs}
\usepackage{graphicx}
\usepackage{subfigure}
\usepackage{tikz}
\usepackage{comment}
\usepackage{amsmath,amssymb} 
\usepackage{color}
\usepackage{multirow}
\usepackage{threeparttable}
\usepackage{booktabs}
\usepackage{ulem}
\begin{document}
\title{A Joint Framework Towards Class-aware and Class-agnostic Alignment for Few-shot Segmentation}
\titlerunning{Joint Class-aware and Class-agnostic Alignment Network}
%
\author{Kai Huang\inst{1}\thanks{Equal contribution.}\orcidID{0000-0003-3700-9052} \and Mingfei Cheng\inst{2}$^{\star}$\orcidID{0000-0002-8982-1483} \and Yang Wang\inst{1} \and Bochen Wang\inst{1} \and Ye Xi\inst{1} \and Feigege Wang\inst{1} \and Peng Chen\inst{1}}
\authorrunning{K. Huang et al.}
%
\institute{Alibaba Group, China\\
\email{\{zhouwan.hk, wanyuan.wy, bochen.wbc, yx150449, feigege.wfgg, yuanshang.cp\}@alibaba-inc.com} \\
\and
Singapore Management University, Singapore\\
\email{mfcheng.2022@phdcs.smu.edu.sg}}
\maketitle              

\begin{abstract}
Few-shot segmentation (FSS) aims to segment objects of unseen classes given only a few annotated support images. Most existing methods simply stitch query features with independent support prototypes and segment the query image by feeding the mixed features to a decoder. Although significant improvements have been achieved, existing methods are still face class biases due to class variants and background confusion. In this paper, we propose a joint framework that combines more valuable class-aware and class-agnostic alignment guidance to facilitate the segmentation. Specifically, we design a hybrid alignment module which establishes multi-scale query-support correspondences to mine the most relevant class-aware information for each query image from the corresponding support features. In addition, we explore utilizing base-classes knowledge to generate class-agnostic prior mask which makes a distinction between real background and foreground by highlighting all object regions, especially those of unseen classes. By jointly aggregating class-aware and class-agnostic alignment guidance, better segmentation performances are obtained on query images. Extensive experiments on PASCAL-$5^i$ and COCO-$20^i$ datasets demonstrate that our proposed joint framework performs better, especially on the 1-shot setting.

\keywords{Few-shot learning \and Semantic segmentation \and Hybrid alignment.}
\end{abstract}
\section{Introduction}
\label{sec:intro}
\begin{figure}[t]
\centering
\setlength{\abovecaptionskip}{0cm}
\setlength{\belowcaptionskip}{0cm}
\subfigure[Conventional Methods]{
    \label{figure1:a}
    \includegraphics[width=5.8cm]{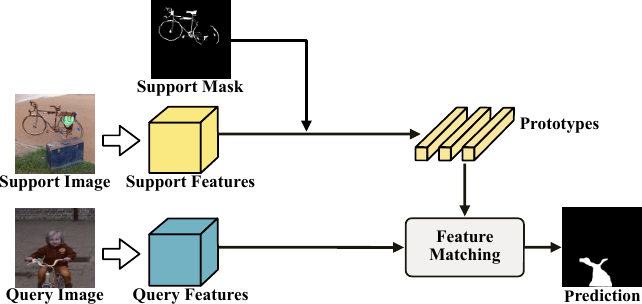}
}
\subfigure[Our Proposed JC$^2$A]{
    \label{figure1:b}
    \includegraphics[width=5.8cm]{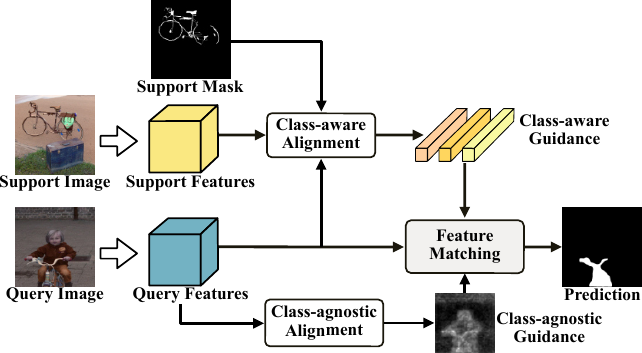}
}
\caption{Illustration of (a) conventional methods and (b) our proposed JC$^2$A. JC$^2$A in (b) joint class-aware and class-agnostic guidance rather than only independent prototypes in (a) to guide the query image segmentation.}
\label{fig:introduction}
\end{figure}

Semantic segmentation has made tremendous progress thanks to the advancement in deep convolutional neural networks. The performance of standard supervised semantic segmentation \cite{DeepLabv3_ECCV_2018,RefineNet_CVPR_2017,Zheng_2021_CVPR} heavily relies on large-scale datasets \cite{VOC2012_2014_IJCV,COCO_ECCV_2014} and will drop drastically on unseen classes. However, obtaining large-scale datasets requires substantial human efforts, which is costly and infeasible in general. Inspired by the few-shot learning \cite{FirstFSS_BMVB_2017}, few-shot segmentation (FSS) has been proposed to alleviate the need of huge annotated data set. Conventional FSS methods are built on meta-learning \cite{snell_2017_NIPS}, which is supposed to learn a generic meta-learner from seen classes and then adopted to handle unseen classes with few annotated support samples. Specifically, as shown in Fig.\ref{fig:introduction}(a), the features of query and support images are firstly extracted by a shared convolutional neural network. Then the support features within the target object regions are transferred to prototypes \cite{Dong_BMVC_2018,Li_CVPR_2021} which are used to guide the query image segmentation with a feature matching module, e.g., relational network \cite{PFENet_TPAMI_2020}.
Despite of the recent progresses made by these work \cite{Li_CVPR_2021,PFENet_TPAMI_2020,Liu_CVPR_2021,wu_ICCV_2021,Mining_ICCV_2021}, there still exist two limitations on FSS. 1) Class-aware bias: The support prototypes extracted independently from the query feature are not discriminative enough due to the variations of objects within the same class. 2) Class-agnostic bias: FSS treats objects of unseen classes as background during training, e.g., \textit{person} in the query image of Fig. \ref{fig:introduction}, which leads to model bias toward the seen classes rather than class-agnostic.

Some researches \cite{Mining_ICCV_2021,BriNet_2020_BMVC,Zhang_ICCV_2019} have tried to address one of the above limitations. To eliminate the influence of class-aware bias, feature interaction \cite{BriNet_2020_BMVC,Zhang_ICCV_2019} are adopted to fuse relative class-aware features in support samples by calculating the pixel-to-pixel relationship between query and support features, which ignore the contextual information and still suffer from the second limitation. Recently, MiningFSS \cite{Mining_ICCV_2021} tries to mine latent object features by using pseudo class-agnostic labels and an extra training branch, which is more complex and does not consider the query-support relationship. To sum up, it is inspiring to study how to fully explore class-aware relationship between query and support samples and class-agnostic information as a joint guidance to improve FSS.

In this work, we propose a novel joint framework, Joint Class-aware and Class-agnostic Alignment Network (JC$^2$A), to address the above-mentioned problems with one stone. For each query image, as shown in Fig. \ref{fig:introduction}(b), JC$^2$A aims to guide the segmentation by jointly aggregating most relevant class-aware information from the support image and class-agnostic information by object region mining. Specifically, JC$^2$A explores class-aware guidance by aligning the prototypes based on the feature relationships between query features and support features, and designs a Hybrid Prototype Alignment Module (HPAM) to build point-to-point and point-to-block correspondences. The class-aware prototypes produced by HPAM contains not only spatial details but also contextual cues of objects in the support feature. To eliminate class-agnostic bias and focus on regions of all objects well, JC$^2$A proposes a Class-agnostic Knowledge Mining Module (CKMM) to mine object regions of all classes in the query image, including seen and unseen classes. The CKMM provides a class-agnostic object mask by highlighting all non-background regions. By aggregating both class-aware prototypes and class-agnostic object mask as a joint guidance, better segmentation performance are obtained on query images. In addition, comprehensive experiments on PASCAL-$5^i$ and COCO-$20^i$ validate the effectiveness of our proposed JC$^2$A in comparison with ablations and alternatives.

\section{Related Work}
\label{sec:relatedwork}
{\bf Semantic Segmentation} is the task to assign a specific category label to each pixel in an image. Inspired by Fully Convolutional Network (FCN) \cite{Long_CVPR_2015}, state-of-the-art segmentation methods \cite{DeepLabv3_ECCV_2018,Zheng_2021_CVPR,Deeplab_TPAMI_2017,PSP_CVPR_2017,Lin_ECCV_2018} have been proposed and applied in various fields \cite{Zheng_2020_CVPR,JTFN_ICCV_2021,Reiss_2021_CVPR,Li_2021_CVPR}. Recently, dilated convolution \cite{DeepLabv3_ECCV_2018,Deeplab_TPAMI_2017,Dilated_ICLR_2016}, pyramid features \cite{PSP_CVPR_2017,PSP2_CVPR_2020}, non-local modules \cite{NonLocal_ICCV_2019,CCNet_ICCV_2019}, vision transformer \cite{Zheng_2021_CVPR} and skip connections \cite{RefineNet_CVPR_2017,UNet_MICCAI_2015} are adopted to perceive more contextual information and preserve spatial details. However, these supervised methods heavily rely on a large amount of pixel-level labeled data. In this work, we focus on FSS which performs better on unseen classes with a handful of annotations.

{\bf Few-shot Learning} is meant to efficient adapt to handle new tasks with limited empirical information available, which emphasizes on the generalization capability of a model. In order to reflect the ability of fitting to new categories given a few annotated data, episodes-based training and verification strategy \cite{vinyals2016matching} has been the foundation of major few-shot learning methods. Meta-based learning methods \cite{finn2017model,lee2019meta,elsken2020meta} maintain a meta-learner to boost the ability of fast acclimatization for new tasks. For instance, meta-manager \cite{finn2017model,lee2018gradient} for parameters optimization, meta-memorizer \cite{ramalho2018adaptive,zhu2018compound} for storing the properties of prototypes and meta-comparator \cite{sung2018learning} for feature retrieval between query image and support set. Metric-based methods \cite{snell_2017_NIPS,karlinsky2019repmet,jiang2020multi} aim to construct a unified similarity measure within the multi-tasks, such as embedding distance of Matching Networks \cite{vinyals2016matching}, parameterized metric of Relation Networks \cite{sung2018learning} and structural distance of DeepEMD \cite{zhang2020deepemd}.

{\bf Few-shot Segmentation} aims to give a dense prediction for the query image with only a few annotated support images. The pioneer work OSLSM \cite{FirstFSS_BMVB_2017} generates segmentation parameters based on support images by the two-branch network including a conditional branch and a segmentation branch. Later, prototype-based methods \cite{Dong_BMVC_2018,Li_CVPR_2021,PFENet_TPAMI_2020,Mining_ICCV_2021,Zhang_CVPR_2021,Xie_CVPR_2021} adopting this two-branch paradigm became the mainstream solutions for FSS. The prototype was first proposed in few-shot segmentation \cite{Dong_BMVC_2018}, which is directly used to guide the segmentation of query images by comparison \cite{Dong_BMVC_2018,PFENet_TPAMI_2020,Mining_ICCV_2021,PANet_ICCV_2019,CANet_CVPR_2019,zhang2021few}. Some works adopt part-aware prototypes \cite{Li_CVPR_2021,yang_ECCV_2020,Liu_ECCV_2020} to contain more diverse support features which may not be needed by the query image. Interaction-based methods \cite{Zhang_ICCV_2019,Wang_ECCV_2020,HSNet_ICCV_2021} extract class-aware information for the query images by only considering point-level correspondence between query and support features. MiningFSS \cite{Mining_ICCV_2021} mines the latent objects by a class-agnostic constraint, which needs extra annotations and parameters. Existing methods rarely focus on both class-aware and class-agnostic information for the query image. In this work, we jointly use class-aware and class-agnostic information as an aggregated guidance for FSS.

\begin{figure}[t]
    \centering
    \setlength{\abovecaptionskip}{0cm}
	\includegraphics[width=\textwidth]{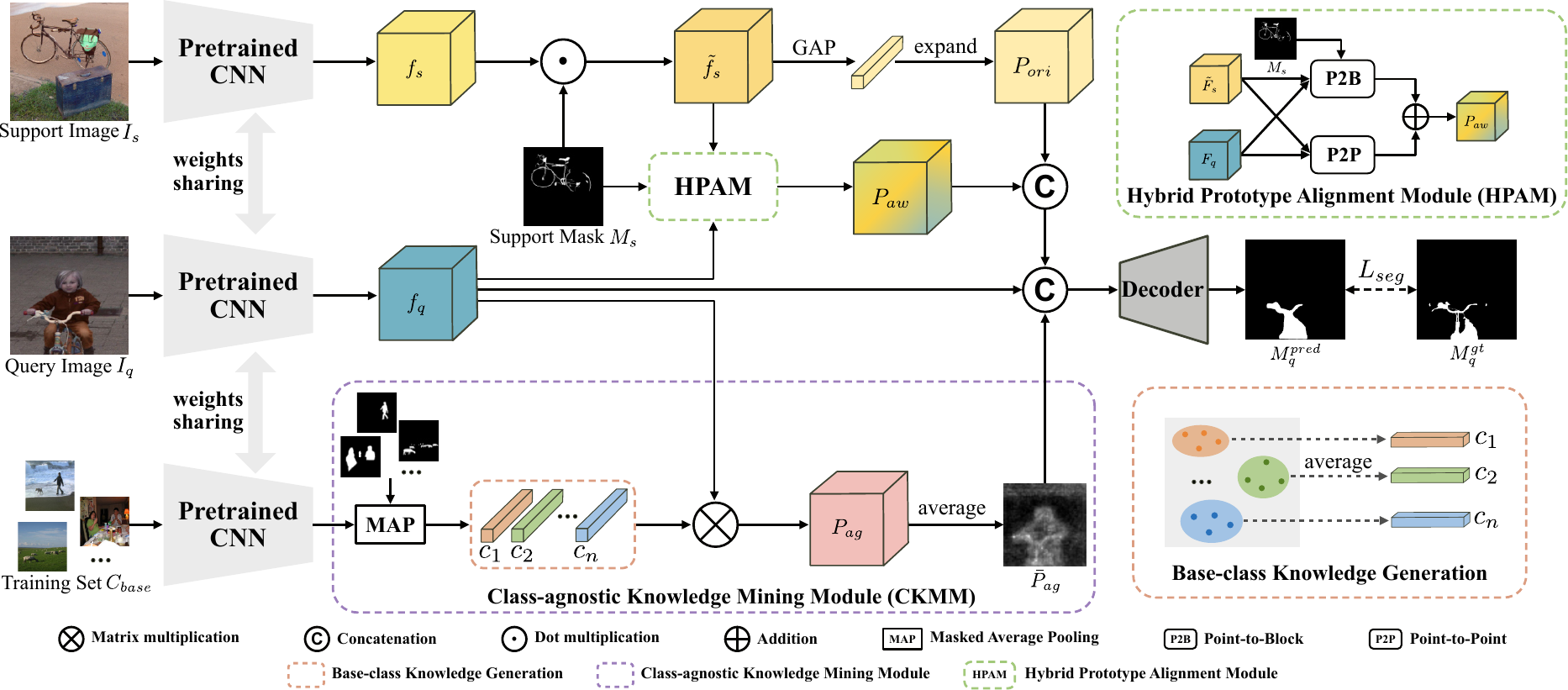}
	\caption{Overview of Joint Class-aware and Class-agnostic Alignment Network (JC$^2$A). A shared encoder is adopted to extract features for support set, query set and base-classes set. Hybrid Prototype Alignment Module (HPAM) and Class-agnostic Knowledge Mining Module (CKMM) are used to generate context-aware prototypes and class-agnostic probability map respectively, which then jointly guide the segmentation.}
 	\label{fig: overview_model}
\end{figure}
\section{Method}
\label{sec:method}
\subsection{Task Formulation}
Few-shot segmentation (FSS) aims to quickly adapt to the given segmentation tasks with only a few annotated data available. For a ${K}$-shot few-shot segmentation task, ${K}$ labeled samples from the same class make up a support set ${S}$ of this task, and there is another unlabeled sample set named query set $Q$. FSS needs to segment the target of the specific class in the query set $Q$ only with the help of the support set ${S}$.

The training and evaluation of FSS models are usually performed by the episodic paradigm \cite{vinyals2016matching}. Specifically, the model is trained on a base-classes set ${C}_{{base}}$ and evaluated on a novel-classes set ${C}_{{novel}}$ with the precondition ${C}_{{base}} \cap {C}_{{novel}} = \emptyset$. For each episode, samples of ${K}$-shot FSS episodic task with class ${c}$ from ${C}_{{base}}$ or ${C}_{{novel}}$ are randomly sampled. The final segmentation performance is reported by averaging the results on the query set of various episodic tasks.

\subsection{Overview}
Fig. \ref{fig: overview_model} illustrates an overview of our joint framework (JC$^2$A) for few-shot segmentation. JC$^2$A mainly contains two components: Hybrid Prototype Alignment Module (HPAM) and Class-agnostic Knowledge Mining Module (CKMM). HPAM is designed to generate most relevant class-aware support prototypes for each query and support image pair by establishing multi-scale query-support relations. Specifically, HPAM builds Point-to-Point and Point-to-Block correspondences between query and support features and combines them as class-aware prototypes, which is able to provide more useful class-aware guidance with spatial details and contextual information. CKMM aims to mitigate the class-agnostic bias mainly caused by background confusion by highlighting all foreground regions in the query image. In CKMM, object features of seen classes in the training set are used to mine the class-agnostic guidance for each query image. More details are introduced in the following.
\subsection{Hybrid Prototype Alignment Module (HPAM)}
Inspired by \cite{Xie_CVPR_2021,Zhang_ICCV_2019}, we utilize the mutual information between the support feature and the query feature to mine more class-aware information from the support feature for the query image. However, simply computing pixel-level attention which contains limited weakened semantics ignores the context. Thus, Hybrid Prototype Alignment Module (HPAM) is proposed to mine the most relevant class-aware guidance from support features by mixing Point-to-Point Alignment and Point-to-Block Alignment, which calculate different scale correspondences between the support and the query.

\textbf{Point-to-Point Alignment (P2P).} Considering the support image $I_s$ and its binary mask $M_s$ as the support instance pair $\{I_s, M_s\}$, and $I_{q}$ denotes the query image, their respective features extracted by the pretrained CNN are represented as $f_{s}$ and $f_{q}$. A Hadamard product is performed in support feature map with its binary mask to obtain the masked support features $\tilde{f}_{s}$. We adopt an attention mechanism to establish the point-to-point relation between $\tilde{f}_{s}$ and $f_{q}$. Formally,
\begin{equation}
   Q_{q} = f_{q}W_{Q}, \quad K_{s} = \tilde{f}_{s}W_{K}, \quad V_{s} = \tilde{f}_{s}W_{V},
    \label{eq: linear_projection}
\end{equation}
where $W_{Q}$, $W_{K}$, $W_{V}$ are learnable projection parameters, $Q_{q}$, $K_{s}$, $V_{s}\in \mathbb R^{C\times (H\times W)}$ are projected features. An attention map is then obtained by the dot-product operation between ${Q_{q}}$ and ${K_{s}}$, which will bring heavy computation as the resolution of feature map grows. To balance computational efficiency and keeping as much target information as possible, a linear attention mechanism \cite{katharopoulos2020transformers,pan2020x} is adopted to decompose the calculation of the attention map, which has linear complexity. Thus, the point-to-point class-aware information can be obtained by the linear attention:
\begin{equation}
     P_{aw}^d = \Phi{(Q_{q})} \times (\Psi{(K_{s})}^{T} \times V_{s}) \in \mathbb R^{C\times (H\times W)},
    \label{eq:DFI_linear_att}
\end{equation}
where $\Phi{(\cdot)}$ and $\Psi{(\cdot)}$ are the decoupling functions to approximate the attention map of normal attention mechanism, in which batch normalization function \cite{ioffe2015batch} and softmax function are commonly used respectively. In order to accommodate the mini-batch learning and suppress the noisy alignment of FSS, we adopt the ReLU function \cite{agarap2018deep} instead of batch normalization to generate positive responses. The Eq. (\ref{eq:DFI_linear_att}) can be rewritten as:
\begin{equation}
    P_{aw}^d = \mathrm{ReLU}(Q_{q}) \times ({softmax}(K_{s})^{T} \times V_{s})
    \in \mathbb R^{C\times (H\times W)}.
    \label{eq: DFI}
\end{equation}
In this way, we obtain the class-aware information of each query point under the full support key-value point features with an dense way. The point-to-point dense alignment between query features and masked support features can offer a detailed and complete pixel level alignment, which tends to find similar grainy vision information from the intra-task targets of support set.

\textbf{Point-to-Block Alignment (P2B).} The Point-to-Point approach only focuses on single point alignment which contains limited weakened semantics. It is observed that the feature points with similar semantic information are always close in spatial locations, consequently, the semantic alignment can be performed in a sparse way with spatial blocked targets of support images. Specifically, We pick out several key feature blocks to represent the class-aware target, such as head-block, leg-block and tail-block for a horse. Then, a point-to-block linear attention alignment between query point-feature and masked support block-feature is formed to catch the semantic class-aware information.

It is worth mentioning that the permutation invariance \cite{lee2019set} of attention mechanism may disrupt the topological relation of feature blocks. Take a horse as an example, it is clear that the head-block is in the front of the tail-block according to the actual spatial location relationship. The topological relation becomes chaotic in the linear attention, which will impact the expression of semantic level information. Thus, we adopt two parameterized matrices to act as the point-specific position embedding, which provides a valid signal which carries the original positional information:
\begin{equation}
     f^{p}_{q} = f_q + p_{q}, 
    \tilde{f}^{p}_{s} = \tilde{f}_s + p_s
\end{equation}
where $p_{q}, p_{s} \in \mathbb R^{H\times W}$ are learnable position embedding for query and support features respectively. Similar to Eq. (\ref{eq: linear_projection}), the queries, keys and values with the position embedding are represented as $Q^{p}_{q}$, $K^{p}_{s}$ and $V^{p}_{s}$. Assume the block area is ${m}\times {m}$ in feature pixels, the concatenated block patch of masked support features is formulated as:
\begin{equation}
    \mathcal{B}(\tilde{f}_{s}^{p}) = {Concat}(\tilde{f}_{s}^{p}[({r}-1){m}:{rm}, ({c}-1){m}:{cm}])_{i=1}^{N}\in \mathbb R^{C\times N\times m^2},
\end{equation}
where ${r}$ is the row position of ${i}$-th block and can be calculated with the round down of ${im}/{W}$, ${c}$ indicates the column position of ${i}$-th block which is obtained with ${i}-{rW}/{m}$, and ${N}={HW}/{m}^2$ is the total number of block patches.

The importance of the ${i}$-th block patch is obtained according to the target coverage with corresponding block in its binary mask:

\begin{equation}
    (\mathcal{I}_{s})_{i} = \sum_{o=({r}-1){m}}^{{rm}}\sum_{j=({c}-1){m}}^{{cm}}M_{s}(o,j)/{m}^{2}.
\end{equation}

The top ${k}$ key feature blocks are selected by the importance ranking within these block patches as:
\begin{equation}
    \mathcal{B}(\tilde{f}_{s}^{p}|{Topk}) = \mathcal{B}(\tilde{f}_{s}^{p}|{Sort}(\mathcal{I}_{s})[:{k}])\in \mathbb R^{{C}\times {k}\times {m}^2}.
\end{equation}
It inevitably contains pure empty regions when the number of valid blocks is smaller than ${k}$. The alignment outcomes of such regions keeps zero response and is insignificant compared with the positive response of those target regions. Therefore, mining an amount of pure empty regions basically does not affect the learning. The {keys} and {values} of blocked support features can be represented as:
\begin{equation}
    \mathcal{B}(K_{s}^{p}), \mathcal{B}(V_{s}^{p}) = \mathcal{B}(\tilde{f}_{s}^{p}W_{S}|{Topk}),\mathcal{B}(\tilde{f}_{s}^{p}W_{V}|{Topk}).
\end{equation}
where the reformulated $\mathcal{B}(K_{s}^{p}), \mathcal{B}(V_{s}^{p})\in \mathbb R^{{C}\times {N}\times {m}^2}$ are the sparsification of support key-value with more focused and explicit semantic information on support features. The sparse feature alignment between query features and blocked support features is further expressed as:
\begin{equation}
\small
    P_{{aw}}^s = \frac{1}{{m}^2}\sum_{{m}\times {m}}\mathrm{ReLU}(Q_{q}^{p}) ({softmax}(\mathcal{B}(K_{s}^{p}))^{T} \mathcal{B}(V_{s}^{p}))\in \mathbb R^{C\times H\times W}.
\small
\end{equation}
Compared with the Point-to-Point prototype alignment in Eq. (\ref{eq: DFI}), we align support prototypes with the blocked support key-value features by using P2B. Moreover, instead of using the block set with a fixed partition, the sparse feature blocks contain pivotal semantic information of targets. Thus the point-to-block prototype alignment aggregates the integrated semantic information with the key feature blocks and offers a block-level semantic alignment, which can generate more stable and smooth class-aware information.

\textbf{Hybrid Prototype.} Our final hybrid aligned prototype combines the aligned prototypes generated by Point-to-Point and Point-to-Block alignment respectively, which is summarized as:

\begin{equation}
    P_{{aw}} = P_{{aw}}^d + P_{{aw}}^s \in \mathbb R^{C\times H\times W}.
\end{equation}

As aforementioned, the P2P alignment is designed to obtain a dense alignment and search the class-aware information with the view of local vision, although the candidate targets of query image usually get positive response, it also introduce background points for the low discrimination of visual features. Thus, by considering the semantic matching between query points and blocked class-aware targets, the P2B alignment can help to filter out these semantic irrelevant points and smooth the class-aware information of P2P alignment.
\textbf{Extension to Feature Pyramid.} Feature pyramid has been widely used in few-shot segmentation \cite{Li_CVPR_2021,PFENet_TPAMI_2020,Zhang_CVPR_2021} due to its abundant multi-scale feature maps. Higher-level hierarchical feature maps possess more centralized semantic information but with lower resolution. In this context, fixed number of key feature blocks become inappropriate, which may introduce noisy information for high-level feature maps or miss essential parts of target for low-level feature maps. In order to tolerate this variation, we apply specific number of block patches corresponding to different scale support feature maps. More concretely, a top ${k}$ set $\{{k}_{{l}}\}_{{l}=1}^{{L}}$ is prepared for ${L}$-layers feature pyramid, ${k}_{{l}}$ decreases as ${l}$ increases. With such a variational top ${k}$ for sparse feature alignment, the blocked support features can gain semantic information of class-aware targets according to the multi-scale feature pyramid.

\subsection{Class-agnostic Knowledge Mining Module (CKMM)}
Due to the limitation of few-shot segmentation datasets, there is only one seen class for the effective targets, and others unseen are treated as the background. The ability of adapting to novel classes is in doubt when same or similar classes are viewed as background in the training process.

Mining latent target with base-classes set was firstly proposed by [53], which focuses on search target by part-specific attributes. However, it suffers from complicated multi-states optimization and the low-semantic target parts are more easy to match the background. Inspired by the feature prototype \cite{snell_2017_NIPS} in few-shot classification, we exploits another simple but effective way with class-specific attitudes. Specifically, we propose to mine the latent target information as well as class-agnostic information with the masked feature prototypes, which are obtained by the base-classes. Given the feature map $f^{c,i}\in \mathbb R^{C\times HW}$ and its binary mask $M^{c,i}$ with class $c$, the spacial weighted Global Average Pooling (wGAP) \cite{Zhang_CVPR_2021,CANet_CVPR_2019} for ${i}$-th instance pair $\{f^{c,i}, M^{c,i}\}$ of class $c$ is defined as:
\begin{equation}
    f_{{wGPA}}^{c,i} = \frac{\sum\nolimits_{h,w}f^{c,i}(h,w)M^{c,i}(h,w)}{\sum\nolimits_{h,w}M^{c,i}(h,w)} \in \mathbb R^{1\times {C}},
\end{equation}
where $h$ and $w$ are the height index and width index with the limitation of $H$ and $W$. Feature prototype of class $c$ is obtained by averaging over the wGAP of all instance pairs with class $c$, and the feature prototype of base-classes $C_{base}$ is concatenated with each single-class feature prototype:

\begin{equation}
\small
    {FP}_{{base}} = {Concat}([\sum\nolimits_{i=1}^{I_{c}}f_{{wGPA}}^{c,i}/I_{c}]_{c \in {C}_{{base}}})
    \in \mathbb R ^ {\lvert {C}_{{base}} \rvert \times {C}},
\small
\end{equation}
where $I_{c}$ is the total instance pairs of class $c$, and $\lvert {C}_{{base}} \rvert$ indicates the cardinality of base-classes set. The feature prototype ${FP}_{{base}}$ can be regarded as the aggregation of seen base-classes, and each row of ${FP}_{{base}}$ contains the most common feature of this class. Subsequently, the latent targets with similar or partially similar features are possibly mined to 
replenish the missing class-agnostic information. The class-agnostic probability map of query image $I_{q}$ is calculated by the dot-product between the base feature prototype and query features:
\begin{equation}
    \bar{P}_{{ag}} = \frac{1}{\lvert {C}_{{base}} \rvert}\sum\nolimits_{\lvert {C}_{{base}} \rvert}{FP}_{{base}}\cdot f_{q}\in \mathbb R^{1\times {W} \times {H}}.
\end{equation}
For the convenience of adapting different base-classes set, we make the average operation in $\lvert {C}_{{base}} \rvert$ dimension to obtain a compositive probability map of class-agnostic information for query features, which acts as a kind of prior mask. Different from the class-specific prior mask in PFENet \cite{PFENet_TPAMI_2020}, our class-agnostic probability map has the ability to search not only the class that is common with support set but also the latent class existed in other meta-tasks.

\subsection{Multiple Information Aggregation (MIA)}
The class-aware information generated by HPAM aims to provide more discriminative prototypes for the current meta-task. The CKMM provides the class-agnostic information to eliminate the background confusion during training. To joint these guidance, we simply combine these two sets of information by concatenating:
\begin{equation}
    P_{{multiple}} = {Concat}([P_{{aw}}, \bar{P}_{{ag}}]) \in \mathbb R^{(C+1)\times H\times W}.
\end{equation}

The concatenated information then is passed through $1\times1$ convolution along with the original query features and support features for further information aggregation. The predicted mask of query image is later obtained by a feature decoder with multi-scale residual layers refer to \cite{PFENet_TPAMI_2020,wu_ICCV_2021}.
\section{Experiments}
\label{sec:ex}
\subsection{Implementation Details}
\label{sec:ID}
{\bf Dataset.} We validate the effectiveness of our proposed method on two standard few-shot segmentation datasets: PASCAL-$5^i$ \cite{FirstFSS_BMVB_2017} and COCO-$20^i$ \cite{Nguyen_ICCV_2019}. {PASCAL-$5^i$} consists of PASCAL VOC 2012 \cite{VOC2012_2014_IJCV} with extra mask annotations from SDS \cite{SDS_ECCV_2014} dataset. It contains 20 object classes which are evenly divided into 4 folds: $\{5^i, i\in\{0,1,2,3\}\}$. {COCO-$20^i$} is a more challenging dataset for few-shot segmentation, which is modified from MS COCO \cite{COCO_ECCV_2014}. It splits 80 categories into 4 folds: $\{20^i, i\in\{0,1,2,3\}\}$. Following the standard experimental settings \cite{PFENet_TPAMI_2020}, on both datasets, three folds are selected for training while the remaining fold is used for evaluation in each single experiment. During the evaluation, 1000 episodes in the target fold are randomly sampled for both datasets.

\noindent {\bf Evaluation metrics.} Following \cite{CANet_CVPR_2019,Nguyen_ICCV_2019}, we adopt mean intersection over union (${m}$IoU) and foreground-background IoU (FB-IoU) as our evaluation metrics. Specifically, mIoU is computed by averaging over IoU values of all classes in a fold. FB-IoU calculates the average of foreground and background IoU in a fold (e.g., $C=2$), which treats all object categories as a single foreground class. The average of all the folds is reported as the final ${m}$IoU/FB-IoU. For the multi-shot case, we leverage the decision-level fusion strategy \cite{Li_CVPR_2021,Zhang_CVPR_2021,CANet_CVPR_2019} by averaging the predicted masks between each single support instance and the query image.

\noindent {\bf Training details.} Our proposed model is constructed on PyTorch \cite{Pytorch_2017_NIPS} and trained on a single NVIDIA RTX 2080Ti. We build our model with the ResNet50 \cite{ResNet_2019_AAAI} and ResNet101 \cite{ResNet_2019_AAAI} as backbones. Our model is optimized by the SGD with an initial learning rate of 2.5$e$-3, where momentum is 0.9 and the weight decay is set to 1$e$-4. During training, the batch size is set to 4 and parameters of the backbone are not updated. All images together with the masks are all resized to $473\times473$ for training and tested with their original sizes. We construct four-layer feature pyramid for the mixed alignment module, the top ${k}$ set is set to \{60, 20, 5, 3\} as $l$ increases, which also corresponds to 20\% number of block patches with respective scales.
\renewcommand{\arraystretch}{1.4}
\begin{table}[t]
   \centering
   \setlength{\abovecaptionskip}{0cm}
   \fontsize{8.5}{8.5}\selectfont
   \caption{Performance of 1-shot and 5-shot segmentation on PASCAL-$5^{i}$. Results in {\bf bold} indicate the best performance and the \uline{underlined} ones are the second best.} 
   \resizebox{\linewidth}{!}{
   \begin{tabular}{ccccccccccccccc}
   \toprule
   \multirow{2}{*}{Backbone} & \multirow{2}{*}{Method} & \multicolumn{6}{c}{1-shot} & \multicolumn{6}{c}{5-shot} \\ \cmidrule(l){3-8} \cmidrule(l){9-14}
   \multicolumn{2}{c}{} & fold-0 & fold-1 & fold-2 & fold-3 & ${m}$IoU & FB-IoU & fold-0 & fold-1 & fold-2 & fold-3 & ${m}$IoU & FB-IoU\cr 
    \midrule
    \multirow{10}{*}{\centering ResNet50}
                                    &PGNet\cite{Zhang_ICCV_2019}& 56.0& 66.9& 50.6& 50.4& 56.0& 69.9& 57.7& 68.7& 52.9& 54.6& 58.5& 70.5\cr
                                    &SCL\cite{Zhang_CVPR_2021}& 63.0& 70.0& 56.5& 57.7& 61.8& 71.9& 64.5& 70.9& 57.3& 58.7& 62.9& 72.8 \cr
                                    &SAGNN\cite{Xie_CVPR_2021}& 64.7& 69.6& 57.0& 57.2& 62.1& 73.2& 64.9& 70.0& 57.0& 59.3& 62.8& 73.3 \cr &CMN\cite{xie_ICCV_2021}&64.3&70.0&57.4&59.4&62.8&72.3&65.8&70.4&57.6&60.8&63.7&72.8\cr
                                    &PFENet\cite{PFENet_TPAMI_2020}& 61.7& 69.5& 55.4& 56.3& 60.8& 73.3& 63.1& 70.7& 55.8& 57.9& 61.9& 73.9\cr
                                    &RePRI\cite{boudiaf_CVPR_2021}& 60.2& 67.0& \uline{61.7} & 47.5& 59.1& -& 64.5& 70.8& \bf 71.7& 60.3& 66.8& - \cr
                                    &MiningFSS\cite{Mining_ICCV_2021}& 59.2& \uline{71.2} & \bf 65.6& 52.5& 62.1& -& 63.5& 71.6& \uline{71.2} & 58.1& 66.1& -\cr 
                                    &HSNet\cite{HSNet_ICCV_2021} & 64.3 & 70.7 & 60.3 & \uline{60.5} & \uline{64.0} & \bf 76.7 & \bf{70.3} & \uline{73.2} & 67.4 & \bf{67.1} & \bf{69.5} & \bf{80.6} \cr
                                    &CyCTR\cite{zhang2021few} & \uline{65.7} & 71.0 & 59.5 & 59.7 & \uline{64.0} & - & \uline{69.3} & \bf 73.5 & 63.8 & \uline{63.5} & \uline{67.5} & - \cr
                                    &\bf JC$^2$A (ours)& \bf 67.3 &\bf 72.4& 57.7&\bf 60.7&\bf 64.5&\uline{76.5}& 68.6& 72.9& 58.7& 62.0& 65.4&\uline{76.8}\cr
                                    
   \midrule
   \multirow{8}{*}{\centering ResNet101}
                                    &PPNet\cite{Liu_ECCV_2020}& 52.7& 62.8& 57.4& 47.7& 55.2& 70.9& 60.3& 70.0& \uline{69.4} & 60.7& 65.1& \uline{77.5}\cr
                                    &DAN\cite{Wang_ECCV_2020}& 54.7& 68.6& 57.8& 51.6& 58.2& 71.9& 57.9& 69.0& 60.1& 54.9& 60.5& 72.3\cr
                                    &PFENet\cite{PFENet_TPAMI_2020}& 60.5& 69.4& 54.4& 55.9& 60.1& 72.9& 62.8& 70.4& 54.9& 57.6& 61.4& 73.5\cr
                                    &RePRI\cite{boudiaf_CVPR_2021}& 59.6& 68.6&\bf 62.2& 47.2& 59.4& -& 66.2& 71.4& 67.0& 57.7& 65.6& -\cr
                                    &MiningFSS\cite{Mining_ICCV_2021}& 60.8& 71.3& 61.5& 56.9& 62.6& -& 65.8& 74.9&\bf 71.4& 63.1&\uline{68.8}& -\cr
                                    &HSNet\cite{HSNet_ICCV_2021} & \uline{67.3} & \uline{72.3} & \uline{62.0} & \bf{63.1} & \uline{66.2} & \uline{77.6} & \bf{71.8} & 74.4 & 67.0 & \bf{68.3} & \bf{70.4} & \bf{80.6} \cr
                                    &CyCTR\cite{zhang2021few} & 67.2 & 71.1 & 57.6 & 59.0 & 63.7 & - & \uline{71.0} & \uline{75.0} & 58.5 & \uline{65.0} & 67.4 & - \cr
                                    &\bf JC$^2$A (ours)&\bf 68.2&\bf 74.4& 59.8&\uline{63.0}&\bf 66.4&\bf 78.8& 70.6&\bf 75.2& 61.9& 64.8& 68.1& \bf 80.6\cr
   \midrule
   \bottomrule            
   \end{tabular}
   }
   \label{tab: pascal_comparison}
\end{table}

\renewcommand{\arraystretch}{1.2}
\begin{table}[h!]
   \centering
   \setlength{\abovecaptionskip}{0cm}
   \fontsize{8.5}{8.5}\selectfont
   \caption{Performance of 1-shot and 5-shot segmentation on COCO-$20^{i}$. Results in {\bf bold} indicate the best performance and the \uline{underlined} ones are the second best.} 
   \resizebox{\linewidth}{!}{
   \begin{tabular}{ccccccccccccccc}
   \toprule
   \multirow{2}{*}{Backbone} & \multirow{2}{*}{Method} & \multicolumn{6}{c}{1-shot} & \multicolumn{6}{c}{5-shot} \\ \cmidrule(l){3-8} \cmidrule(l){9-14}
   \multicolumn{2}{c}{} & fold-0 & fold-1 & fold-2 & fold-3 & ${m}$IoU & FB-IoU & fold-0 & fold-1 & fold-2 & fold-3 & ${m}$IoU & FB-IoU\cr 
   \midrule
    \multirow{8}{*}{\centering ResNet50}
                                    &PPNet\cite{Liu_ECCV_2020}& 31.5 & 22.6 &21.5&16.2&23.0&-& \uline{45.9} &29.2&30.6&29.6&33.8&-\cr &RePRI\cite{boudiaf_CVPR_2021}&31.2&38.1&33.3&33.0&34.0&-&38.5&46.2&40.0&43.6&42.1&-\cr &MMNet\cite{wu_ICCV_2021}&34.9&41.0&37.2&37.0&37.5&-&37.0&40.3&39.3&36.0&38.2&-\cr &CMN\cite{xie_ICCV_2021}&37.9&\uline{44.8}&38.7&35.6&39.3&61.7&42.0&50.5&41.0&38.9&43.1&63.3\cr
                                    &CyCTR\cite{zhang2021few}& 38.9 & 43.0 & \uline{39.6} & \uline{39.8} & \uline{40.3} & - & 41.1 & 48.9 & 45.2 & \bf{47.0} & 45.6 & -\cr
                                    
                                    &MiningFSS\cite{Mining_ICCV_2021}&\bf 46.8&35.3&26.2&27.1&33.9&-&\bf 54.1&41.2&34.1&33.1&40.6&-\cr 
                                    &HSNet\cite{HSNet_ICCV_2021}&36.3&43.1&38.7&38.7&39.2& \uline{68.2} & 43.3 & \uline{51.3} & \bf{48.2} & 45.0 & \uline{46.9} & \uline{70.7}\cr 
                                    
                                    &\bf JC$^2$A (ours)&\uline{40.4}&\bf 47.4&\bf 44.5&\bf 43.5&\bf 44.0&\bf 70.0&44.3&\bf 53.5& \uline{46.0} & \uline{45.8} &\bf 47.4&\bf 71.5\cr

   \midrule
   \multirow{7}{*}{\centering ResNet101}
                                    &PMMs\cite{yang_ECCV_2020}&29.5&36.8&28.9&27.0&30.6&-&33.8&42.0&33.0&33.3&35.5&-\cr
                                    &PFENet\cite{PFENet_TPAMI_2020}& 34.3& 33.0& 32.3& 30.1& 32.4& 58.6& 38.5& 38.6& 38.2& 34.3& 37.4& 61.9\cr  &SCL\cite{Zhang_CVPR_2021}&36.4&38.6&37.5&35.4&37.0&-&38.9&40.5&41.5&38.7&39.9&-\cr &SAGNN\cite{Xie_CVPR_2021}&36.1&41.0&38.2&33.5&37.2&60.9&40.9&48.3&42.6&38.9&42.7&63.4\cr
                                    &MiningFSS\cite{Mining_ICCV_2021}&\bf 50.2&37.8&27.1&30.4&36.4&-&\bf 57.0&46.2&37.3&37.2&44.4&-\cr 
                                    &HSNet\cite{HSNet_ICCV_2021}& 37.2 & \uline{44.1} & \uline{42.4} & \uline{41.3} & \uline{41.2} & \uline{69.1} & \uline{45.9} & \uline{53.0} & \bf{51.8} & \uline{47.1} & \bf{49.5} & \bf{72.4} \cr
                                    &\bf JC$^2$A (ours)& \uline{41.5} & \bf 48.6&\bf 45.6&\bf 42.9&\bf 44.7&\bf 70.6&43.7&\bf 55.2& \uline{47.3} &\bf 47.7& \uline{48.5} & \uline{72.0} \cr
   \midrule
   \bottomrule            
   \end{tabular}
   }
   \label{tab: coco_comparison}
\end{table}
\subsection{Comparisons}
To verify the effectiveness of our proposed method, we compare with alternatives on the two few-shot segmentation datasets \cite{COCO_ECCV_2014,FirstFSS_BMVB_2017}. Extensive experiments with various backbones show that our model achieves the best performance as shown in Table \ref{tab: pascal_comparison} and Table \ref{tab: coco_comparison}.

\noindent {\bf Quantitative results.} In Table \ref{tab: pascal_comparison}, we show the comparative results of our JC$^2$A and alternative FSS methods on PASCAL-$5^i$. Although our method does not perform better on PASCAL-$5^i$ 5-shot, our method achieves competitive performance compared with other methods on the 1 shot setting. The highest increment $m$IoU based metrics is around 2 points (e.g. from 72.3\% to 74.4\% for fold-1 with ResNet101). Table \ref{tab: coco_comparison} presents the results of different approaches on COCO-20$^i$. It can be found that our JC$^2$A outperforms significantly compared with alternatives on both 1-shot and 5-shot settings. With the backbone of ResNet50, our method outperforms the second best by 3.7\% ${m}$IoU and 0.5\% ${m}$IoU on 1-shot setting and 5-shot setting respectively. The performance gains with different backbones further demonstrate the superiority of our JC$^2$A, particularly with the backbone of ResNet101 on COCO-20$^i$, which exceeds the second best model by 3.5\% on 1-shot. From the above comparison, we conclude that our JC$^2$A achieves better performance. Besides, we think that JC$^2$A is more suitable for few-shot segmentation, because it obtains the SOTA on 1-shot setting, which means fewer annotated samples are required in JC$^2$A.

\begin{figure}[t]
    \centering
    \setlength{\abovecaptionskip}{0cm}
    \includegraphics[width=0.9\textwidth]{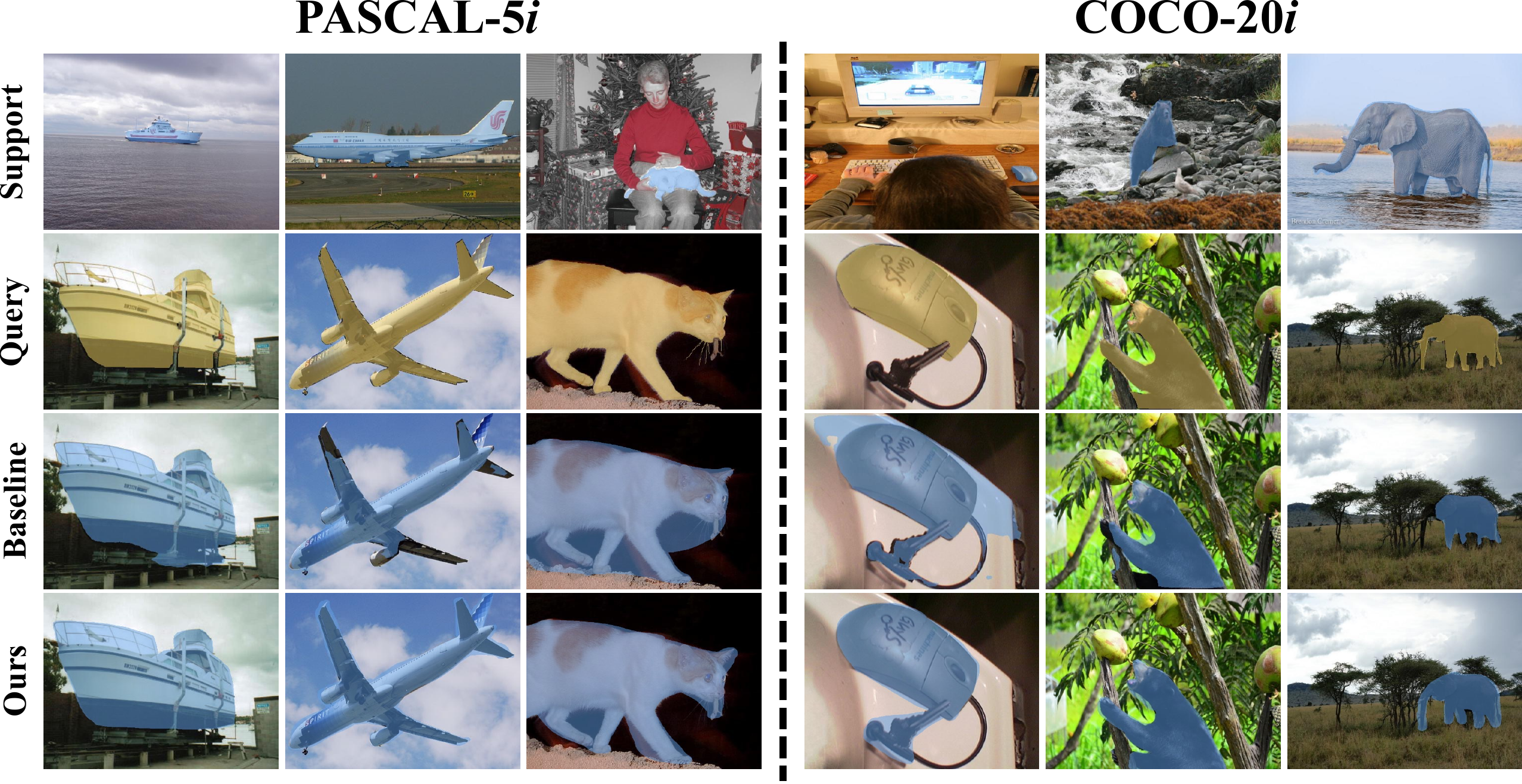}
    \caption{Qualitative results on PASCAL-$5^{i}$ and COCO-$20^i$. Oriented top to bottom, each row shows the ground truth of query images (\textcolor[RGB]{238,206,99}{yellow}), the baseline results (\textcolor[RGB]{95,161,233}{blue}) and ours results (\textcolor[RGB]{95,161,233}{blue}), respectively.}
    \label{fig: qualitative_results}
\end{figure}

\noindent {\bf Qualitative results.} Fig \ref{fig: qualitative_results} provides visual examples of JC$^2$A on PASCAL-$5^i$ and COCO-20$^i$. Compared our results (the 4th row) with the baseline (the 3rd row), JC$^2$A yields fewer false predictions in base classes and background. Besides, JC$^2$A can capture more details and maintain a more complete structure of the target object. These results verify that the joint class-aware and class-agnostic guidance is effective for FSS.

\renewcommand{\arraystretch}{1.0} 
\begin{table}[h]  
  \centering  
  \setlength{\abovecaptionskip}{0cm}
  \caption{Ablation Study on the effect of different components. {``P2P''} and {``P2B''} represent the Point-to-Point alignment and the Point-to-Block alignment respectively.}
  \begin{threeparttable}
  \setlength{\tabcolsep}{2mm}{
  \begin{tabular}{cccccc}
  \toprule
  \multirow{2}{*}{P2P} & \multirow{2}{*}{P2B} & \multirow{2}{*}{CKMM} & \multirow{2}{*}{Parameters} & \multicolumn{2}{c}{$m$IoU} 
  \\ \cmidrule(l){5-6}
  \multicolumn{4}{c}{} & 1-shot & 5-shot \cr
  \midrule
  & & & 34.09M & 59.0 & 60.6 \cr
  \checkmark& & & 34.52M & 61.3 & 62.2 \cr
  & \checkmark & & 34.61M & 62.7 & 63.3  \cr
  \checkmark & \checkmark & & 34.61M & 63.6 & 64.7  \cr
  & & \checkmark & 34.11M & 61.7 & 62.8 \cr
  \checkmark & \checkmark & \checkmark & 34.64M & \bf 64.5 & \bf 65.4  \cr
  \bottomrule  
  \end{tabular}}  
  \end{threeparttable} 
   \label{tab: ablation_study_component}
\end{table}

\renewcommand{\arraystretch}{1.0} 
\begin{table}[h]  
  \centering  
  \caption{Ablation Study on HPAM. ``NA'' is the normal attention, ``NLA'' indicates the normal linear attention, ``PE'' means the position embedding in Point-to-Block alignment, ``IS'' is the inference speed on 1-shot setting.}
  \begin{threeparttable}
  \setlength{\tabcolsep}{2mm}{
  \begin{tabular}{ccccccc}
  \toprule
  \multirow{2}{*}{Setting} & \multicolumn{3}{c}{1-shot} & \multicolumn{3}{c}{5-shot} \\ \cmidrule(l){2-4} \cmidrule(l){5-7}
  \multicolumn{1}{c}{} & ${m}$IoU $\uparrow$ & FB-IoU $\uparrow$ & IS $\uparrow$ & ${m}$IoU $\uparrow$ & FB-IoU $\uparrow$ & IS $\uparrow$ \cr
  \midrule
  NA & 64.2 & 76.9 & 1.00x & 65.5 & 77.2 & 0.19x\cr
  NLA & 62.8 & 74.5 & 4.11x & 63.3 & 74.9 & 0.84x\cr
  cosine & 62.3 & 74.4 & 2.86x & 62.8 & 73.3 & 0.55x\cr
  Ours {w}/{o} PE & 63.1 & 74.9 & 4.07x & 64.0 & 75.2 & 0.80x\cr
  Ours & 64.5 & 76.5 & 4.10x & 65.4 & 76.8 & 0.82x\cr
  \bottomrule  
  \end{tabular}}
  \end{threeparttable} 
   \label{tab: ablation_study_mixed_feature_alignment}
\end{table}
\renewcommand{\arraystretch}{1.4}
\begin{table}[h]
   \centering
   \setlength{\abovecaptionskip}{0cm}
   \fontsize{8.5}{8.5}\selectfont
   \caption{Effectiveness of CKMM (Class-agnostic guidance) for different encoder-decoder based FSS methods on PASCAL-$5^{i}$ and COCO-20$^i$.} 
   \resizebox{\linewidth}{!}{
   \begin{tabular}{ccccccccccccc}
   \toprule
   \multirow{2}{*}{Dataset} & \multirow{2}{*}{Method} & \multicolumn{5}{c}{1-shot} & \multicolumn{5}{c}{5-shot} \\ \cmidrule(l){3-7} \cmidrule(l){8-12}
   \multicolumn{2}{c}{} & fold-0 & fold-1 & fold-2 & fold-3 & \textit{m}IoU & fold-0 & fold-1 & fold-2 & fold-3 & \textit{m}IoU\cr 
   \midrule
    \multirow{3}{*}{\centering PASCAL-5$^i$}
                                    &PFENet\cite{PFENet_TPAMI_2020}& 61.7\tiny{+0.82}& 69.5\tiny{+0.59}& 55.4\tiny{+1.24}& 56.3\tiny{+1.05}& 60.8\tiny{+0.93}& 63.1\tiny{+0.65}& 70.7\tiny{+1.14}& 55.8\tiny{+0.98}& 57.9\tiny{+0.77}& 61.9\tiny{+0.89}\cr
                                    &SCL\cite{Zhang_CVPR_2021}& 63.0\tiny{+0.65}& 70.0\tiny{+1.63}& 56.5\tiny{+0.47}& 57.7\tiny{+0.70}& 61.8\tiny{+0.86}& 64.5\tiny{+0.79}& 70.9\tiny{+1.28}& 57.3\tiny{+0.54}& 58.7\tiny{+0.77}& 62.9\tiny{+0.85} \cr
                                    &MM-Net\cite{wu_ICCV_2021}& 62.7\tiny{+1.72}& 70.2\tiny{+0.60}& 57.3\tiny{+0.54}& 57.0\tiny{+0.98}& 61.8\tiny{+0.96}& 62.2\tiny{+1.89}& 71.5\tiny{+1.05}& 57.5\tiny{+0.66}& 62.4\tiny{+0.96}& 63.4\tiny{+1.14} \cr 
                                    
   \midrule
   \multirow{3}{*}{\centering COCO-20$^i$}
                                    &PFENet\cite{PFENet_TPAMI_2020}& 34.3\tiny{+1.05}& 33.0\tiny{+0.89}& 32.3\tiny{+0.67}& 30.1\tiny{+0.90}& 32.4\tiny{+0.88}& 38.5\tiny{+0.94}& 38.6\tiny{+0.77}& 38.2\tiny{+0.93}& 34.3\tiny{+0.84}& 37.4\tiny{+0.87}\cr
                                    &SCL\cite{Zhang_CVPR_2021}& 36.4\tiny{+1.26}& 38.6\tiny{+1.30}& 37.5\tiny{+0.78}& 35.4\tiny{+1.03}& 37.0\tiny{+1.09}& 38.9\tiny{+1.17}& 40.5\tiny{+1.34}& 41.5\tiny{+0.88}& 38.7\tiny{+1.01}& 39.9\tiny{+1.10} \cr
                                    &MM-Net\cite{wu_ICCV_2021}& 35.4\tiny{+1.51}& 41.7\tiny{+0.90}& 37.5\tiny{+1.33}& 40.1\tiny{+1.06}& 36.2\tiny{+1.20}& 37.8\tiny{+1.66}& 41.0\tiny{+1.11}& 40.3\tiny{+1.28}& 36.9\tiny{+1.37}& 39.0\tiny{+1.36}\cr
   \midrule
   \bottomrule            
   \end{tabular}
   }
   \label{tab: CKMM_comparison}
\end{table}

\subsection{Ablation Studies}
To analyze the impact of each component in JC$^2$A, we conduct extensive ablation studies on PASCAL-$5^i$. Here, our baseline model is obtained from JC$^2$A excluding HPAM and CKMM.

\noindent \textbf{Model Effectiveness.} We first conduct an ablation study to show the effectiveness of the Hybrid Prototype Alignment Module (HPAM) and Class-agnostic Knowledge Mining Module (CKMM). Results are summarized in Table \ref{tab: ablation_study_component}. It is noted that the model using HPAM (P2P$+$P2B) outperforms the baseline (1st row) by 4.6\% and 4.1\% on 1-shot and 5-shot settings respectively. CKMM provides class-agnostic information to FSS by highlighting all object regions. Observing results of 1st row and 5th row in Table \ref{tab: ablation_study_component}, we can see CKMM improves the results by a large margin with 2.7\% $m$IoU on 1-shot and 2.2\% $m$IoU on 5-shot, which shows the effectiveness of CKMM. The last row of Table \ref{tab: ablation_study_component} demonstrates that the combination of these two modules performs better than only using each of them. We can infer that HPAM and CKMM mutually benefit during meta-learning.
\begin{figure}[t]
    \centering
    \includegraphics[width=0.9\textwidth]{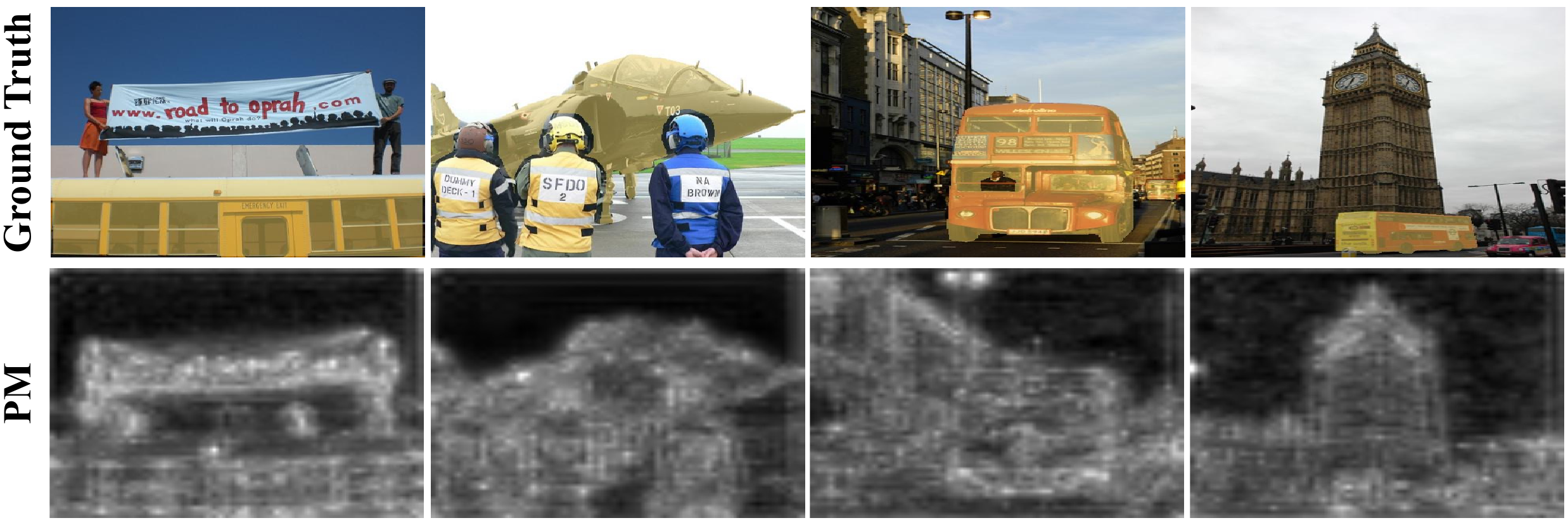}
    \setlength{\abovecaptionskip}{0cm}
    \caption{Visualization of class-agnostic probability maps (PM) generated by CKMM. The 1st row shows query images with their annotations (\textcolor[RGB]{238,206,99}{yellow}). The 2nd row shows probability maps which highlight all object regions of seen and unseen classes.}
    \label{fig: Visualization_probability_map}
\end{figure}

\noindent \textbf{Hybrid Prototype Alignment Module (HPAM).} HPAM contains different scale prototype alignments, P2P and P2B. The 2nd row and the 3rd row in Table \ref{tab: ablation_study_component} proves the effectiveness of combination of P2P and P2B. Table \ref{tab: ablation_study_mixed_feature_alignment} studies the influence of operations in HPAM. Since the feature alignment is accomplished by a modified linear attention, we compare it with the normal attention (NA), normal linear attention (NLA) and cosine interaction. It is clear that our class- aware feature alignment achieves competitive performance with greatly increased inference speed. Although the normal attention way gets slight superiority in some cases, it trades off with a huge computational cost which is reflected in its slowest inference speed. For the reason of mini-batch data form in few-shot segmentation, the normalization decoupling function suffers from instability of data distribution and gets worse performance than ReLU function adopted in our method. The interaction ability of cosine similarity is weaker than the attention-based measure for its lack of nonlinear mapping and noisy suppression. Besides, as shown in the 4th row and the bottom row of Table \ref{tab: ablation_study_mixed_feature_alignment}, the position embedding also plays a positive role in our method to improve the results.

\noindent \textbf{Class-agnostic Knowledge Mining Module (CKMM).} CKMM is designed to provide class-agnostic alignment guidance for FSS by highlighting all object regions. As shown in Fig \ref{fig: Visualization_probability_map}, CKMM is able to successfully highlight all object regions. To further demonstrate the effectiveness of CKMM and its generated class-agnostic probability maps (PM), we apply it to several encoder-decoder based FSS methods \cite{PFENet_TPAMI_2020,wu_ICCV_2021,Zhang_CVPR_2021}. We adopt ResNet50 and ResNet101 as the backbone of PASAL-5$^i$ and COCO-20$^i$ dataset respectively. The experimental results in Table \ref{tab: CKMM_comparison} indicate that our CKMM can also boost other FSS approaches without upsetting the original structures. It also proves that the class-agnostic alignment guidance is beneficial for FSS.

\noindent \textbf{Hybrid Prototypes \& Information Aggregation.} Table \ref{tab: way_aggregate_information} shows the ablation study of different information aggregation methods. For the aggregation of obtaining hybrid prototypes from P2P and P2B, the \textit{add} operation obtains better performance than others. The recommended operation is \textit{concat} between mixed alignment for class-specific targets and probability map for class-agnostic targets. It is reasonable that the \textit{add} operation is more suitable to aggregate the information with similar properties and the \textit{concat} operation prefers differentiated information with the precondition of absence of curse of dimensionality.
\renewcommand{\arraystretch}{1.0} 
\begin{table}[t]  
  \centering  
  \caption{Ablation study on different ways of hybrid prototypes and information aggregation. Three common operations are compared: Multiply, Add and Concat.}
  \begin{threeparttable}
  \setlength{\tabcolsep}{2mm}{
  \begin{tabular}{cccccc}
  \toprule
  \multirow{2}{*}{Component} & \multirow{2}{*}{Setting} & \multicolumn{2}{c}{1-shot} & \multicolumn{2}{c}{5-shot} \\ \cmidrule(l){3-4} \cmidrule(l){5-6}
  \multicolumn{2}{c}{} & ${m}$IoU & FB-IoU & ${m}$IoU & FB-IoU\cr
  \midrule
  \multicolumn{1}{c}{\multirow{3}{*}{\begin{tabular}[c]{@{}c@{}}HPAM\end{tabular}}}
  & \textit{Multiply} & 60.2 & 72.3 & 61.8 & 73.5\cr
  \multicolumn{1}{c}{} &{\bf \textit{Add}} &\bf 64.5 &\bf 76.5 &\bf 65.4 &\bf 76.8\cr
  \multicolumn{1}{c}{} & \textit{Concat} & 64.0 & 75.7 & 64.8 & 76.1\cr
  \midrule
  \multicolumn{1}{c}{\multirow{3}{*}{\begin{tabular}[c]{@{}c@{}}MIA\end{tabular}}}
  & \textit{Multiply} & 58.8 & 70.1 & 60.9 & 73.0\cr
  \multicolumn{1}{c}{} & \textit{Add} & 62.7 & 74.3 & 63.3 & 75.0\cr
  \multicolumn{1}{c}{} &{\bf \textit{Concat}} &\bf 64.5 &\bf 76.5 &\bf 65.4 &\bf 76.8\cr
  \bottomrule  
  \end{tabular}}
  \end{threeparttable} 
   \label{tab: way_aggregate_information}
\end{table}
\section{Conclusion}
\label{sec:conclusion}
In this paper, we have proposed a joint framework JC$^2$A towards class-aware and class-agnostic alignment for few-shot segmentation. JC$^2$A contains two critical modules: Hybrid Prototype Alignment Module (HPAM) and Class-agnostic Knowledge Mining Module (CKMM), then combines these two modules to jointly guide the query image segmentation. HPAM aims to generate class-aware guidance for the query image by combining multi-scale aligned prototypes between query features and support features. To prevent background confusion and class-agnostic bias, CKMM uses base-classes knowledge to produce a class-agnostic probability mask for the query image, which highlights object regions of all classes especially those of unseen classes. Comparisons with FSS alternatives validate the effectiveness of joint class-aware and class-agnostic information in guiding the query image segmentation. Potential extensions of JC$^2$A include developing more replaceable components for each module, thus improving FSS performance.
%
%
%
\bibliographystyle{splncs04}
\bibliography{egbib}

\begin{thebibliography}{10}
\providecommand{\url}[1]{\texttt{#1}}
\providecommand{\urlprefix}{URL }
\providecommand{\doi}[1]{https://doi.org/#1}

\bibitem{agarap2018deep}
Agarap, A.F.: Deep learning using rectified linear units (relu). arXiv preprint
  arXiv:1803.08375  (2018)

\bibitem{boudiaf_CVPR_2021}
Boudiaf, M., Kervadec, H., Masud, Z.I., Piantanida, P., Ben~Ayed, I., Dolz, J.:
  Few-shot segmentation without meta-learning: A good transductive inference is
  all you need? In: CVPR (2021)

\bibitem{Deeplab_TPAMI_2017}
Chen, L.C., Papandreou, G., Kokkinos, I., Murphy, K., Yuille, A.L.: Deeplab:
  Semantic image segmentation with deep convolutional nets, atrous convolution,
  and fully connected crfs. TPAMI  \textbf{40} (2017)

\bibitem{DeepLabv3_ECCV_2018}
Chen, L.C., Zhu, Y., Papandreou, G., Schroff, F., Adam, H.: Encoder-decoder
  with atrous separable convolution for semantic image segmentation. In: ECCV
  (2018)

\bibitem{JTFN_ICCV_2021}
Cheng, M., Zhao, K., Guo, X., Xu, Y., Guo, J.: Joint topology-preserving and
  feature-refinement network for curvilinear structure segmentation. In: ICCV
  (2021)

\bibitem{Dong_BMVC_2018}
Dong, N., Xing, E.P.: Few-shot semantic segmentation with prototype learning.
  In: BMVC (2018)

\bibitem{elsken2020meta}
Elsken, T., Staffler, B., Metzen, J.H., Hutter, F.: Meta-learning of neural
  architectures for few-shot learning. In: CVPR (2020)

\bibitem{VOC2012_2014_IJCV}
Everingham, M., Eslami, S.M.A., Gool, L.V., Williams, C.K.I., Winn, J.M.,
  Zisserman, A.: The pascal visual object classes challenge: A retrospective.
  International Journal of Computer Vision  \textbf{111} (2015)

\bibitem{finn2017model}
Finn, C., Abbeel, P., Levine, S.: Model-agnostic meta-learning for fast
  adaptation of deep networks. In: ICML. PMLR (2017)

\bibitem{SDS_ECCV_2014}
Hariharan, B., Arbelaez, P., Girshick, R.B., Malik, J.: Simultaneous detection
  and segmentation. In: ECCV (2014)

\bibitem{ResNet_2019_AAAI}
Hu, T., Yang, P., Zhang, C., Yu, G., Mu, Y., Snoek, C.G.M.: Attention-based
  multi-context guiding for few-shot semantic segmentation. In: AAAI (2019)

\bibitem{CCNet_ICCV_2019}
Huang, Z., Wang, X., Huang, L., Huang, C., Wei, Y., Liu, W.: Ccnet: Criss-cross
  attention for semantic segmentation. In: ICCV (2019)

\bibitem{ioffe2015batch}
Ioffe, S., Szegedy, C.: Batch normalization: Accelerating deep network training
  by reducing internal covariate shift. In: ICML (2015)

\bibitem{jiang2020multi}
Jiang, W., Huang, K., Geng, J., Deng, X.: Multi-scale metric learning for
  few-shot learning. IEEE Transactions on Circuits and Systems for Video
  Technology  \textbf{31}(3),  1091--1102 (2020)

\bibitem{karlinsky2019repmet}
Karlinsky, L., Shtok, J., Harary, S., Schwartz, E., Aides, A., Feris, R.,
  Giryes, R., Bronstein, A.M.: Repmet: Representative-based metric learning for
  classification and few-shot object detection. In: CVPR (2019)

\bibitem{katharopoulos2020transformers}
Katharopoulos, A., Vyas, A., Pappas, N., Fleuret, F.: Transformers are rnns:
  Fast autoregressive transformers with linear attention. In: ICML (2020)

\bibitem{lee2019set}
Lee, J., Lee, Y., Kim, J., Kosiorek, A., Choi, S., Teh, Y.W.: Set transformer:
  A framework for attention-based permutation-invariant neural networks. In:
  ICML (2019)

\bibitem{lee2019meta}
Lee, K., Maji, S., Ravichandran, A., Soatto, S.: Meta-learning with
  differentiable convex optimization. In: CVPR (2019)

\bibitem{lee2018gradient}
Lee, Y., Choi, S.: Gradient-based meta-learning with learned layerwise metric
  and subspace. In: ICML. PMLR (2018)

\bibitem{Li_CVPR_2021}
Li, G., Jampani, V., Sevilla-Lara, L., Sun, D., Kim, J., Kim, J.: Adaptive
  prototype learning and allocation for few-shot segmentation. In: CVPR (2021)

\bibitem{Li_2021_CVPR}
Li, Y., Zhao, H., Qi, X., Wang, L., Li, Z., Sun, J., Jia, J.: Fully
  convolutional networks for panoptic segmentation. In: CVPR (2021)

\bibitem{Lin_ECCV_2018}
Lin, D., Ji, Y., Lischinski, D., Cohen-Or, D., Huang, H.: Multi-scale context
  intertwining for semantic segmentation. In: ECCV (2018)

\bibitem{RefineNet_CVPR_2017}
Lin, G., Milan, A., Shen, C., Reid, I.: Refinenet: Multi-path refinement
  networks for high-resolution semantic segmentation. In: CVPR (2017)

\bibitem{COCO_ECCV_2014}
Lin, T., Maire, M., Belongie, S.J., Bourdev, L.D., Girshick, R.B., Hays, J.,
  Perona, P., Ramanan, D., Doll{\'{a}}r, P., Zitnick, C.L.: Microsoft {COCO:}
  common objects in context. In: ECCV (2014)

\bibitem{Liu_CVPR_2021}
Liu, B., Ding, Y., Jiao, J., Ji, X., Ye, Q.: Anti-aliasing semantic
  reconstruction for few-shot semantic segmentation. In: CVPR (2021)

\bibitem{Liu_ECCV_2020}
Liu, Y., Zhang, X., Zhang, S., He, X.: Part-aware prototype network for
  few-shot semantic segmentation. In: ECCV (2020)

\bibitem{Long_CVPR_2015}
Long, J., Shelhamer, E., Darrell, T.: Fully convolutional networks for semantic
  segmentation. In: CVPR (2015)

\bibitem{HSNet_ICCV_2021}
Min, J., Kang, D., Cho, M.: Hypercorrelation squeeze for few-shot segmentation.
  In: ICCV (2021)

\bibitem{Nguyen_ICCV_2019}
Nguyen, K., Todorovic, S.: Feature weighting and boosting for few-shot
  segmentation. In: ICCV (2019)

\bibitem{pan2020x}
Pan, Y., Yao, T., Li, Y., Mei, T.: X-linear attention networks for image
  captioning. In: CVPR (2020)

\bibitem{Pytorch_2017_NIPS}
Paszke, A., Gross, S., Chintala, S., Chanan, G., Yang, E., DeVito, Z., Lin, Z.,
  Desmaison, A., Antiga, L., Lerer, A.: Automatic differentiation in pytorch.
  In: NeurIPS Autodiff Workshop (2017)

\bibitem{ramalho2018adaptive}
Ramalho, T., Garnelo, M.: Adaptive posterior learning: few-shot learning with a
  surprise-based memory module. In: ICLR (2018)

\bibitem{Reiss_2021_CVPR}
Reiss, S., Seibold, C., Freytag, A., Rodner, E., Stiefelhagen, R.: Every
  annotation counts: Multi-label deep supervision for medical image
  segmentation. In: CVPR (2021)

\bibitem{UNet_MICCAI_2015}
Ronneberger, O., Fischer, P., Brox, T.: U-net: Convolutional networks for
  biomedical image segmentation. In: MICCAI (2015)

\bibitem{FirstFSS_BMVB_2017}
Shaban, A., Bansal, S., Liu, Z., Essa, I., Boots, B.: One-shot learning for
  semantic segmentation. In: BMVC (2017)

\bibitem{snell_2017_NIPS}
Snell, J., Swersky, K., Zemel, R.: Prototypical networks for few-shot learning.
  In: NeurIPS (2017)

\bibitem{sung2018learning}
Sung, F., Yang, Y., Zhang, L., Xiang, T., Torr, P.H., Hospedales, T.M.:
  Learning to compare: Relation network for few-shot learning. In: CVPR (2018)

\bibitem{PFENet_TPAMI_2020}
Tian, Z., Zhao, H., Shu, M., Yang, Z., Li, R., Jia, J.: Prior guided feature
  enrichment network for few-shot segmentation. TPAMI  (2020)

\bibitem{vinyals2016matching}
Vinyals, O., Blundell, C., Lillicrap, T., Wierstra, D., et~al.: Matching
  networks for one shot learning. In: NIPS (2016)

\bibitem{Wang_ECCV_2020}
Wang, H., Zhang, X., Hu, Y., Yang, Y., Cao, X., Zhen, X.: Few-shot semantic
  segmentation with democratic attention networks. In: ECCV (2020)

\bibitem{PANet_ICCV_2019}
Wang, K., Liew, J.H., Zou, Y., Zhou, D., Feng, J.: Panet: Few-shot image
  semantic segmentation with prototype alignment. In: ICCV (2019)

\bibitem{wu_ICCV_2021}
Wu, Z., Shi, X., Lin, G., Cai, J.: Learning meta-class memory for few-shot
  semantic segmentation. In: Proceedings of the IEEE/CVF International
  Conference on Computer Vision. pp. 517--526 (2021)

\bibitem{Xie_CVPR_2021}
Xie, G.S., Liu, J., Xiong, H., Shao, L.: Scale-aware graph neural network for
  few-shot semantic segmentation. In: CVPR (2021)

\bibitem{xie_ICCV_2021}
Xie, G.S., Xiong, H., Liu, J., Yao, Y., Shao, L.: Few-shot semantic
  segmentation with cyclic memory network. In: ICCV (2021)

\bibitem{yang_ECCV_2020}
Yang, B., Liu, C., Li, B., Jiao, J., Ye, Q.: Prototype mixture models for
  few-shot semantic segmentation. In: ECCV (2020)

\bibitem{Mining_ICCV_2021}
Yang, L., Zhuo, W., Qi, L., Shi, Y., Gao, Y.: Mining latent classes for
  few-shot segmentation. In: ICCV (2021)

\bibitem{BriNet_2020_BMVC}
Yang, X., Wang, B., Chen, K., Zhou, X., Yi, S., Ouyang, W., Zhou, L.: Brinet:
  Towards bridging the intra-class and inter-class gaps in one-shot
  segmentation. In: BMVC (2020)

\bibitem{Dilated_ICLR_2016}
Yu, F., Koltun, V.: Multi-scale context aggregation by dilated convolutions.
  In: ICLR (2016)

\bibitem{Zhang_CVPR_2021}
Zhang, B., Xiao, J., Qin, T.: Self-guided and cross-guided learning for
  few-shot segmentation. In: CVPR (2021)

\bibitem{zhang2020deepemd}
Zhang, C., Cai, Y., Lin, G., Shen, C.: Deepemd: Few-shot image classification
  with differentiable earth mover's distance and structured classifiers. In:
  CVPR (2020)

\bibitem{Zhang_ICCV_2019}
Zhang, C., Lin, G., Liu, F., Guo, J., Wu, Q., Yao, R.: Pyramid graph networks
  with connection attentions for region-based one-shot semantic segmentation.
  In: ICCV (2019)

\bibitem{CANet_CVPR_2019}
Zhang, C., Lin, G., Liu, F., Yao, R., Shen, C.: Canet: Class-agnostic
  segmentation networks with iterative refinement and attentive few-shot
  learning. In: CVPR (2019)

\bibitem{zhang2021few}
Zhang, G., Kang, G., Yang, Y., Wei, Y.: Few-shot segmentation via
  cycle-consistent transformer. In: NIPS (2021)

\bibitem{PSP_CVPR_2017}
Zhao, H., Shi, J., Qi, X., Wang, X., Jia, J.: Pyramid scene parsing network.
  In: CVPR (2017)

\bibitem{PSP2_CVPR_2020}
Zhen, M., Wang, J., Zhou, L., Li, S., Shen, T., Shang, J., Fang, T., Quan, L.:
  Joint semantic segmentation and boundary detection using iterative pyramid
  contexts. In: CVPR (2020)

\bibitem{Zheng_2021_CVPR}
Zheng, S., Lu, J., Zhao, H., Zhu, X., Luo, Z., Wang, Y., Fu, Y., Feng, J.,
  Xiang, T., Torr, P.H., Zhang, L.: Rethinking semantic segmentation from a
  sequence-to-sequence perspective with transformers. In: CVPR (2021)

\bibitem{Zheng_2020_CVPR}
Zheng, Z., Zhong, Y., Wang, J., Ma, A.: Foreground-aware relation network for
  geospatial object segmentation in high spatial resolution remote sensing
  imagery. In: CVPR (2020)

\bibitem{zhu2018compound}
Zhu, L., Yang, Y.: Compound memory networks for few-shot video classification.
  In: ECCV (2018)

\bibitem{NonLocal_ICCV_2019}
Zhu, Z., Xu, M., Bai, S., Huang, T., Bai, X.: Asymmetric non-local neural
  networks for semantic segmentation. In: ICCV (2019)

\end{thebibliography}
%




\end{document}